\documentclass[letterpaper]{article} 
\usepackage{aaai2026}  
\usepackage{times}  
\usepackage{helvet}  
\usepackage{courier}  
\usepackage[hyphens]{url}  
\usepackage{graphicx} 
\usepackage{amsmath} 
\usepackage{natbib}  
\usepackage{caption} 
\usepackage{algorithm}
\usepackage{algorithmic}

\usepackage{newfloat}
\usepackage{listings}
\DeclareCaptionStyle{ruled}{labelfont=normalfont,labelsep=colon,strut=off} 
\floatstyle{ruled}
\newfloat{listing}{tb}{lst}{}
\floatname{listing}{Listing}
\title{Stochasticity in Agentic Evaluations: Quantifying Inconsistency with Intraclass Correlation}
\author{
    Zairah Mustahsan,
    Abel Lim,
    Megna Anand,
    Saahil Jain,
    Bryan McCann
}
\affiliations{
    zairah@you.com, 
    abel@you.com,
    megna@you.com, 
    saahil@you.com, 
    bryan@you.com
}

\usepackage{bibentry}

\begin{document}

\maketitle

\begin{abstract}
As large language models become components of larger agentic systems, evaluation reliability becomes critical: unreliable sub-agents introduce brittleness into downstream system behavior. Yet current evaluation practice, reporting a single accuracy number from a single run, obscures the variance underlying these results, making it impossible to distinguish genuine capability improvements from lucky sampling.
We propose adopting Intraclass Correlation Coefficient (ICC), a metric from measurement science, to characterize this variance. ICC decomposes observed variance into between-query variance (task difficulty) and within-query variance (agent inconsistency), highlighting whether reported results reflect true capability or measurement noise.
We evaluated on GAIA (Levels 1–3, measuring agentic capabilities across varying reasoning complexity) and FRAMES (measuring retrieval and factuality across multiple documents). We found that ICC varies dramatically with task structure, with reasoning and retrieval tasks (FRAMES) exhibit ICC=0.4955–0.7118 across models, and agentic tasks (GAIA) exhibiting ICC= 0.304-0.774 across models. For sub-agent replacement decisions in agentic systems, accuracy improvements are only trustworthy if ICC also improves. We demonstrate that ICC converges by n=8–16 trials for structured tasks and $n\geq 32$ for complex reasoning, enabling practitioners to set evidence-based resampling budgets. We recommend reporting accuracy alongside ICC and within-query variance as standard practice, and propose updated Evaluation Cards capturing these metrics. By making evaluation stability visible, we aim to transform agentic benchmarking from opaque leaderboard competition to trustworthy experimental science. Our code is open-sourced at \url{ https://github.com/youdotcom-oss/stochastic-agent-evals}. 
\end{abstract}


\section{Introduction}

Large language models (LLMs) are no longer confined to static text prediction. 
Increasingly, they act as agents: using tools, interacting with environments, and carrying out multi-step plans. 
To measure such capabilities, the field has turned to agentic evaluations \citet{gaia2024},  \citet{zhang2024dseval}, and \citet{xu2023toolbench}. 
These benchmarks are rapidly becoming reference points for progress. 
Yet the way they are currently used reduces complex stochastic processes to a single leaderboard number. 
That number, typically an accuracy or success rate from one run, obscures the variance that determines whether the result 
is reproducible at all \citet{miller2024statsevals} \citet{li2025arenalite}.

Agentic evaluations should be approached as experiments, with outcomes analyzed for variability and reproducibility. 
Reported performance is often based on a single trial of an inherently random process, with little visibility into uncertainty. 
Sources of randomness include sampling inside the language model, the behavior of external APIs, and the design of the scoring 
function. Without accounting for these factors, comparisons across agents risk overstating differences and underestimating variability.

We examine the stochasticity of agentic evaluations using GAIA and FRAMES as running case studies. Replicating these benchmark runs across multiple trials and all three difficulty levels shows that evaluation stability varies significantly with task complexity and model capability, a finding obscured by current single-run reporting practices. Our contribution is both diagnostic and prescriptive: we diagnose why current evaluations are fragile, and we propose ICC (intraclass correlation coefficient) as a metric practitioners can use to characterize and report on evaluation stability, enabling more principled experimental design for agentic benchmarks.

\section{Background and Motivation}

Agentic evaluations differ in structure from static NLP benchmarks. Whereas traditional benchmarks test a model’s ability to map an input to a single output, agentic settings involve multi-step decision making, tool use, and interactions with dynamic environments. An agent may need to query an API, manipulate files, or perform calculations before arriving at an answer. These differences make agentic evaluations richer, but also more sensitive to randomness.

Current practice has yet to reflect this added complexity. Many benchmarks report results from a single run per agent, without confidence intervals, standard errors, or replication. Some studies attempt to reduce bias from agent inconsistency by reporting average@k values (where k=3), but still do so without providing confidence intervals or statistical significance measures (\citet{yao2024taubenchbenchmarktoolagentuserinteraction}\citet{tongyidr}). Comparisons between agents are frequently made without statistical testing, and scoring functions are not always defined in detail. As a result, observed differences may partly reflect random variation, scorer choices, or agent inconsistency rather than genuine differences in capability.

Recent work has begun to address this gap. \citet{miller2024statsevals} provides a statistical treatment of LLM evaluations, deriving estimators and error formulas and recommending resampling protocols. Platforms such as Chatbot Arena and Arena-Lite have introduced bootstrapped 95\% confidence 
intervals into leaderboards \citet{li2025arenalite}, highlighting that uncertainty can be made visible at scale. 
\citet{blackwell2024uncertainty} argue for explicit uncertainty quantification and reproducible protocols in LLM benchmarking, 
while \citet{bowyer2025central} caution against naive reliance on the Central Limit Theorem in small benchmarks due to under-coverage. 
More broadly, surveys of evaluation methodologies for LLM-based agents emphasize the lack of standardized protocols and reproducibility practices \citet{yehudai2025survey}.

The intraclass correlation coefficient (ICC) \citep{shrout1979intraclass, mcgraw1996forming} has been widely used in psychometrics and medical research to assess measurement reliability by decomposing variance into between-subject and within-subject components. \citet{koo2016guideline} provide guidelines for selecting and interpreting ICC variants across different study designs. While ICC is standard practice for evaluating inter-rater reliability and test-retest consistency in clinical settings, it has not been systematically applied to agentic AI evaluation, where variance arises from both task difficulty and agent stochasticity.

Beyond statistical treatments, several strands of prior work contextualize our analysis. Large-scale benchmarks such as BIG-bench \citet{srivastava2022bigbench} and \citet{liang2022helm} established multi-dimensional evaluation but did not incorporate statistical testing. 
MT-Bench \citet{zheng2023mtbench} popularized human–LLM comparison, again without replication. Infrastructure projects such as the 
lm-eval-harness \citet{gao2021lmevalharness}, OpenAI Evals \citet{chen2023openaievals}, and Dynabench \citet{kiela2021dynabench} emphasize standardization and robustness, 
though they rarely quantify uncertainty. Parallel work on uncertainty and calibration 
\citet{kadavath2022know}, \citet{chiang2024chatbotarena}, and \citet{liu2023uncertainty} highlights confidence 
estimation at the token or output level rather than full-agent evaluation. 
Together, this literature underlines both the momentum and the methodological gaps that motivate our study.

\section{Stochasticity in Agentic Evaluations}
\begin{figure}[H]
\centering
  \includegraphics[width=0.45\textwidth]{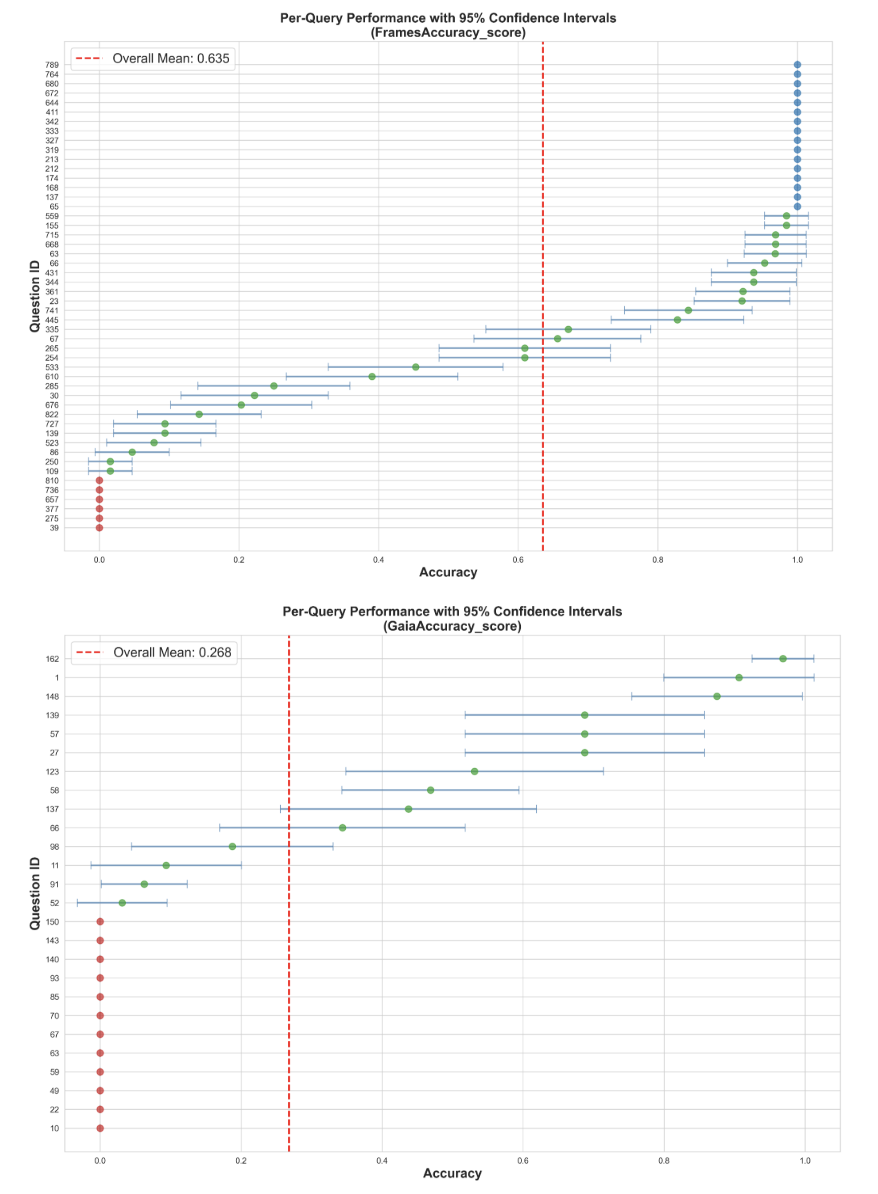}
    \caption{Per-question accuracy with 95\% confidence intervals across sampled questions, 64 trials per question. FRAMES and GAIA with GPT-4o search. Wide confidence intervals show trial-to-trial variance in agent behavior.}
    \label{fig:question_variance_gaia}
\end{figure}

Agentic evaluations are inherently stochastic processes. The underlying LLM models sample from probability distributions rather than deterministically computing outputs \cite{brown2020language, holtzman2020curious, he2025defeating}, introducing trial-to-trial variance.. This inherent stochasticity is compounded by the evaluation framework: task specifications, API configuration, retry policies, timeout behavior, and external environments all introduce trial-to-trial variance that ICC directly measures.
Each major component contributes to within-query variance: task specifications may be ambiguous; agents sample stochastically or encounter tool errors; environments introduce latency or rate limits; evaluation frameworks govern retry and timeout behavior; scoring functions rely on heuristics such as fuzzy matching or numeric normalization; and the choice of trial count determines whether results reflect reliable averages or noisy draws. As frameworks such as the Model Context Protocol (MCP) \citet{anthropic2023mcp} become standard for tool integration, server configuration and versioning introduce additional trial-level variability. Table \ref{tab:anatomy} summarizes these components and their sources of within-query variance: the trial-to-trial inconsistency that ICC directly measures. Most benchmarks do not quantify them explicitly. Figure ~\ref{fig:question_variance_gaia} illustrates this variance empirically: per-question accuracy estimates with 95\% confidence intervals show substantial trial-to-trial inconsistency that single-run reporting obscures.

\begin{table*}[t]
\centering
\resizebox{0.95\textwidth}{!}{
\begin{tabular}{l|l|l|l}
    \textbf{Component} & \textbf{Role} & \textbf{Source of variance} & \textbf{Quantified?} \\
    \hline
    Task specification & Defines the goal and tool usage & Ambiguity, underspecification & Rarely \\
    Agent (LLM + plugins) & Executes decisions and sampling actions & Model decoding randomness, plugin errors & No \\
    Environment & Provides external dynamics or APIs & Latency, rate limits, unstable responses & No \\
    Evaluation framework & Orchestrates interaction loop & Seeds, retries, exponential backoff, timeouts & No \\
    Scoring function & Determines correctness & Fuzzy matching, normalization heuristics & Sometimes \\
\end{tabular}
}
\caption{Components of an agentic evaluation and their typical sources of variance.}
\label{tab:anatomy}
\end{table*}

We formalize evaluation as a stochastic function where each trial produces a potentially different outcome:
\\
  \begin{equation}
  \text{EvalScore}_i(t) = f(\text{Task}_i, \text{Agent}, \text{Env}, \text{Scorer})
  \label{eq:evalscore}
  \end{equation}
  \\
\text{where } $i$ \text{ indexes questions and } $t \in \{1, \ldots, T\}$ indexes independent runs, the Agent includes both the model and its toolchain, the Env encompasses all external APIs or simulated world dynamics, the Scorer encodes normalization and comparison rules, and variation across t arises from the model's stochastic sampling.\\

  \subsection{Recommended Statistical Protocols for Agentic Benchmarks}

  \subsubsection{Variance Estimation and Confidence Intervals}

Any reported point estimate without a measure of variance is incomplete. Agentic evaluations are inherently stochastic, and a single accuracy number obscures the underlying variability. To estimate the mean performance of an agent across tasks, we conduct multiple independent runs—each with a different random outcome. The variance of this estimator depends on both the number of tasks and the number of trials per task.

Let \( A^{(t)}_i \in \{0,1\} \) denote correctness of output \( i \) in trial \( t \), for \( n \) items and \( T \) trials. The overall accuracy estimate is:
  
\begin{equation}
\hat{\mu} = \frac{1}{Tn} \sum_{t=1}^T \sum_{i=1}^n A^{(t)}_i .
\label{eq:mean}
\end{equation}

Using the standard error of the mean,
  
\begin{equation}
SE(\hat{\mu}) = \sqrt{ \frac{\hat{\mu}\,(1 - \hat{\mu})}{Tn} } ,
\label{eq:stderr}
\end{equation}
  
we can construct \( (1-\alpha) \) confidence intervals:
    
\begin{equation}
\hat{\mu} \pm z_{\alpha/2}\, SE(\hat{\mu}) ,
\label{eq:ci}
\end{equation}

where \( z_{\alpha/2} \) is the critical value of the normal distribution. Reporting confidence intervals enables readers to assess whether observed differences between agents are likely to be meaningful or attributable to noise. The variance of this estimator declines with \( T \); thus, even a modest replication (e.g., \( T \geq 5 \)) substantially stabilizes estimates.

\subsubsection{Testing Agent Differences with Dependent Samples}
  
  When comparing two agents on the same evaluation items, predictions are paired.
  We therefore recommend methods that exploit this pairing. McNemar’s test
  \citet{dietterich1998approximate} computes a $p$-value from
  discordant outcomes \( n_{01} \) and \( n_{10} \) (cases where one agent is correct
  and the other is not):
  
  \begin{equation}
  \chi^2 = \frac{(|n_{01} - n_{10}| - 1)^2}{n_{01} + n_{10}} ,
  \label{eq:mcnemar}
  \end{equation}

  which approximately follows a chi-squared distribution with 1 degree of freedom.
  Alternatively, paired bootstrap methods \citep{efron1994introduction}
  resample items with replacement to generate confidence intervals for the difference
  in accuracy.

\subsubsection{Variance Decomposition and Intraclass Correlation Coefficient (ICC)}

When evaluating agents across multiple trials, performance varies due to two distinct sources: (1) some tasks are inherently harder than others (between-task variance), and (2) the agent behaves inconsistently on the same task across trials (within-task variance). Understanding this decomposition is essential for evaluation stability and forms the foundation for using the intraclass correlation coefficient (ICC) as a reliability metric.
For full mathematical framework see Appendix E.

\subsubsection{Intraclass Correlation Coefficient}

The ICC quantifies the proportion of total variance attributable to differences between tasks and can be understood as a function of both the difficulty of the dataset (variance between tasks) and the consistency of the agent (variance within tasks).

\begin{equation}
\text{ICC} = \frac{\sigma_b^2}{\sigma_b^2 + \sigma_w^2}
\label{eq:icc}
\end{equation}

We employ the one-way random effects model, ICC(1,1) \citep{shrout1979intraclass}, which assumes:
\begin{enumerate}
    \item Tasks are random effects sampled from a larger population of possible evaluation items
    \item Trials are random effects representing stochasticity in the agent's behavior
    \item We report single-trial reliability: ICC(1,1) measures the reliability of an individual trial, with within-task variance estimated by pooling variances across all tasks
\end{enumerate}

This model treats both the selection of tasks and the variability across trials as sources of random variation, making it appropriate when evaluation tasks represent a sample from a broader domain of interest. The between-task variance $\sigma_b^2$ is estimated from the variance of task means:

\begin{equation}
\sigma_b^2 = \frac{1}{n-1} \sum_{i=1}^{n} (\bar{Y}_{i\cdot} - \bar{Y}_{\cdot\cdot})^2
\label{eq:between_variance}
\end{equation}

where $\bar{Y}_{i\cdot}$ is the mean performance on task $i$ and $\bar{Y}_{\cdot\cdot}$ is the grand mean across all tasks and trials. Note that when computing variance estimates from multiple trials per task, we use task-level means rather than pooling individual trial scores, as the latter would violate independence assumptions \citep{miller2024statsevals}.

\subsubsection{ICC Estimation and Sample Size}

In practice, we compute an estimated ICC, denoted $\widehat{\text{ICC}}$, based on $n$ tasks and $T$ trials per task. Both sample sizes affect the precision of this estimate.

\paragraph{Effect of Sample Size}

The estimated ICC converges to the true population ICC as sample sizes increase:

\begin{itemize}
    \item \textbf{More tasks ($n$)}: Improves the estimate of between-task variance $\hat{\sigma}_b^2$. With few tasks, we may not capture the full range of task difficulties.
    
    \item \textbf{More trials per task ($T$)}: Improves the estimate of within-task variance $\hat{\sigma}_w^2$. The contribution of within-task variance to total variance scales as $\sigma_w^2/T$ \citep{miller2024statsevals}, so more trials reduce its impact.
\end{itemize}

The precision of $\widehat{\text{ICC}}$ can be quantified by its standard error, which for ICC(1,1) is approximately \citep{bonettintraclass2002}:

\begin{equation}
\text{SE}(\widehat{\text{ICC}}) \approx \sqrt{\frac{2(1-\widehat{\text{ICC}})^2(1+(T-1)\widehat{\text{ICC}})^2}{n(n-1)(T-1)(F^2)}}
\label{eq:icc_se}
\end{equation}
where \(F\) is the F-statistic from a one-way repeated measures ANOVA.
This decreases with both $n$ and $T$, with the relative benefit of additional trials depending on the true ICC value.

\paragraph{Practical Guidelines for better ICC estimates}

When reporting ICC values:
\begin{itemize}
    \item Aim for $SE(ICC) \leq 0.03$, 
  which yields a 95\% confidence interval of ±0.06 around 
  your ICC estimate. This means if you report ICC = 0.65, 
  you can be confident the true ICC lies between 0.59–0.71.
    \item Report both $n$ and $T$ alongside ICC values
    \item With low $T$, ICC estimates are less reliable. Increasing $T$ provides more confidence in the measured ICC value
\end{itemize}

\subsubsection{Interpretation of ICC}

ICC values provide standardized measures of evaluation stability \citep{koo2016guideline}:

\begin{itemize}
    \item \textbf{ICC $\geq$ 0.75}: Good reliability. Task difficulty dominates variance. Agent responses are consistent across trials. Most observed variance reflects true differences in task difficulty.
    \item \textbf{ICC = 0.50--0.75}: Moderate reliability. Mixed contribution from task difficulty and agent stochasticity. Both sources matter for evaluation outcomes.
    \item \textbf{ICC $<$ 0.50}: Poor reliability. Agent is highly inconsistent, or tasks lack sufficient difficulty variation. Difficult to isolate true performance differences.
\end{itemize}

\textbf{Why ICC complements accuracy reporting:} Consider two evaluations where an agent achieves 73\% accuracy:

\begin{center}
\begin{tabular}{c|c|p{4cm}}
    \textbf{Accuracy} & \textbf{ICC} & \textbf{Interpretation} \\
    \hline
    73\% & 0.66 & Tasks separate by difficulty. Agent behavior is consistent across trials. \\
    73\% & 0.30 & Agent behavior is unpredictable. Same task produces different results across trials. \\
\end{tabular}
\end{center}

Both agents achieve 73\% accuracy, but the first is far more consistent. ICC provides insight into agent reliability beyond raw performance scores. When agents have similar accuracy, ICC serves as a tiebreaker—higher ICC indicates more predictable behavior. ICC also helps distinguish genuine capability limitations from measurement noise, informing resampling strategies and benchmark design.

\subsubsection{Important Caveats: When Lower ICC Is Expected}
Note: Lower ICC does not necessarily indicate a ”bad”
benchmark. Some tasks may inherently require exploration
or have legitimate stochasticity in their solutions. However,
low ICC does signal that:
\begin{enumerate}
    \item Single-run results are unreliable.
    \item Practitioners must understand what variance reflects agent uncertainty vs. task ambiguity.
    \item Heavier resampling is required for trustworthy conclusions.
\end{enumerate}
\subsubsection{Running more trials vs sampling more tasks}
In practice, dataset subsampling is often applied during evaluations due to limited computational budgets. Since variance arises from two sources: \emph{between-item} variance (\(\sigma^2_b\), items differ in difficulty) and \emph{within-item} variance (\(\sigma^2_w\), trial-to-trial randomness). The variance decomposes as:

\begin{equation}
\mathrm{Var}(\hat{\mu}) = \frac{\sigma^2_b}{n} + \frac{\sigma^2_w}{nT} .
\label{eq:var_decomposed}
\end{equation}

For a fixed computational budget \( B = nT \), we can rewrite this as:

\begin{equation}
\mathrm{Var}(\hat{\mu}) = \frac{\sigma^2_b}{n} + \frac{\sigma^2_w}{B} .
\label{eq:var_budget}
\end{equation}

The second term is constant for fixed \(B\); only the first term depends on our allocation choice. Therefore, variance is minimized by \textbf{maximizing} \(n\) (i.e., evaluating more items with fewer trials each). For example, with \(B = 400\) and typical variance ratios \(\sigma^2_b / \sigma^2_w \approx 5\), allocating \(n=100, T=4\) yields 68\% lower standard error than \(n=10, T=40\) (see Figure~\ref{fig:variance_comparison} in Appendix A).

This strategy is optimal until all available items are exhausted (\(n = n_{\max}\)), at which point increasing \(T\) is the only option for further variance reduction.

\section{Datasets and Evaluation Scope}

\textbf{GAIA Benchmark}: GAIA's \citet{gaia2024} validation set consists of roughly 160 questions across three difficulty levels, each requiring reasoning, calculation, or tool use. GAIA's design emphasizes short, unambiguous answers, typically numbers or brief strings, enabling automated scoring. Crucially, reported results in the original paper present single-run accuracy values with no discussion of trial-to-trial variability, uncertainty, or consistency. This makes GAIA representative of current evaluation practice: ambitious in scope, widely cited, but methodologically opaque regarding reliability. GAIA levels are defined as: Level 1 (no tools or single tool, $\leq5$ steps, 53 questions), Level 2 (5–10 steps, multiple tools, 86 questions), Level 3 (arbitrary action sequences, unrestricted tools, 26 questions).

\textbf{FRAMES Benchmark}: FRAMES (Factuality, Retrieval, And reasoning Measurement Set)  \citet{krishna2024frames} is a dataset of 824 test samples designed to evaluate LLMs' ability to retrieve and reason across multiple documents in end-to-end RAG (retrieval-augmented generation) scenarios. Unlike GAIA's open-ended reasoning, FRAMES tasks focus on information retrieval grounded in factuality—the agent must retrieve and integrate correct information across multiple sources. FRAMES comprises challenging multi-hop questions requiring integration of information from multiple sources, with baseline LLM performance around 0.408 accuracy.

\textbf{Why these two}: GAIA and FRAMES represent diversity in task structure and complexity. GAIA's open-ended reasoning with multi-modality naturally introduces higher trial-to-trial variance, while FRAMES' focused retrieval-and-reasoning scope may exhibit different variance patterns. Together, they illustrate how task design, not just model capability, predicts evaluation stability.

\subsection{Evaluation Scope and Setup}

\begin{enumerate}
    \item Dataset sample: For GAIA, we evaluated on the validation set across all three difficulty levels. For FRAMES, a random sample of 50 questions $(random\_state=42)$ was used due to computational cost concerns.
    \item Number of trials: We ran 64 trials per question on both datasets to provide a large sample for analyzing trial-to-trial variance across task difficulty levels. For o4-mini-deep research, we ran 8 trials per question due to computational cost concerns.
    \item Prompting strategy: We used a standard agentic reasoning prompt (see Appendix B). 
    \item Tool availability: We evaluated agents with and without web search or deep research 
   capabilities via OpenAI, Claude, Gemini APIs. We also experimented with open-source models like Qwen and DeepSeek. Model specifications and API identifiers 
   are provided in Appendix C.
    \item Scoring: We used o4-mini as a LLM judge to evaluate whether agent outputs matched ground-truth answers. 
    \item Failures handling: Timeouts were set to 120 seconds per query. Failed runs (timeouts, unrecoverable errors) were recorded as incorrect answers, reflecting real-world deployment scenarios.
\end{enumerate}

\section{Experiments}

\subsection{GAIA: ICC in Multi-Modal Reasoning and Tool-Use Tasks}

We evaluated GPT-4o search and GPT-5 search across GAIA Levels 1, 2, and 3, running 64 trials per question. Table \ref{tab:icc_by_level} shows results.

\begin{table*}[!h]
\centering
\small
\begin{tabular}{c|cc|cc|cc|cc}
    \textbf{Level} & \multicolumn{2}{c|}{\textbf{Accuracy}} & \multicolumn{2}{c|}{\textbf{95\% CI}} & \multicolumn{2}{c|}{\textbf{Between Var}} & \multicolumn{2}{c}{\textbf{ICC}} \\
    & \textbf{GPT-4o} & \textbf{GPT-5} & \textbf{GPT-4o} & \textbf{GPT-5} & \textbf{GPT-4o} & \textbf{GPT-5} & \textbf{GPT-4o} & \textbf{GPT-5} \\
    \hline
    Level 1 (53 Q) & 22.7\% & 62.3\% & [14.0\%, 31.4\%] & [52.9\%, 71.7\%] & 0.100 & 0.185 & \textbf{0.561} & \textbf{0.774} \\
    Level 2 (86 Q) & 23.2\% & 54.2\% & [15.8\%, 30.6\%] & [44.9\%, 63.5\%] & 0.119 & 0.187 & \textbf{0.662} & \textbf{0.745} \\
    Level 3 (26 Q) & 6.6\% & 44.2\% & [1.0\%, 12.2\%] & [28.1\%, 60.4\%] & 0.019 & 0.160 & \textbf{0.304} & \textbf{0.629} \\
    \hline
\end{tabular}
\begin{center}
{\tiny $^*$Both GPT-4o and GPT-5 were evaluated with web search enabled.}
\end{center}
\caption{ICC and variance decomposition across GAIA levels (full validation set) and models (64 trials per question).}
\label{tab:icc_by_level}
\end{table*}

\begin{table*}[!h]
\centering
\small
\begin{tabular}{c|c|c|c|c}
    \textbf{Model} & {\textbf{Accuracy}} & {\textbf{95\% CI}} & {\textbf{Between Var}} & {\textbf{ICC}} \\
    \hline
    GPT-5 search & 77.31\% & [68.86\%, 85.77\%] & 0.088 & \textbf{0.496} \\
    GPT-4o search & 63.54\% & [51.70\%, 75.38\%] & 0.174 & \textbf{0.735} \\
    GPT-4o & 38.16\% & [26.40\%, 49.91\%] &0.171 & \textbf{0.712} \\
    
 Claude 4.5 Haiku & 68.37\% & [57.58\%, 79.17\%] & 0.144 & \textbf{0.655}\\
 Claude 4.5 Sonnet & 66.44\% & [55.20\%, 77.68\%] & 0.156 & \textbf{0.689}\\

Gemini 2.5 Pro & 62.34\% & [50.60\%, 74.09\%] & 0.174 & \textbf{0.713}\\
  Qwen3-235b-a22b & 34.22\% & [23.53\%, 44.91\%] & 0.169 & \textbf{0.617}\\
 Deepseek-v3p1 & 44.75\% & [33.13\%, 56.37\%] & 0.157 & \textbf{0.663}\\
\end{tabular}
\begin{center}
{\tiny $^*$GPT-5 \& Claude family were evaluated with web search, GPT-4o with and without web search and others without web search.}
\end{center}
\caption{ICC and variance decomposition on FRAMES (n=50, 64 trials per question).}
\label{tab:frames_table}
\end{table*}

\subsubsection{Interpretation by Level} 

~

\textbf{Levels 1–2 (Easier-Medium Reasoning):}
Both levels exhibit moderate-to-high ICC for GPT-4o (0.561–0.662) and high ICC for GPT-5 (0.745–0.774), indicating questions clearly separate by difficulty and agent behavior is relatively consistent. GPT-5 outperforms GPT-4o substantially: Level 1 (+39.6 pp accuracy, +0.213 ICC) and Level 2 (+31.0 pp accuracy, +0.083 ICC). The combined accuracy and ICC improvements indicate GPT-5 is not just more capable but also more reliable—performance gains reflect genuine capability gains, not fortunate sampling. \\
\textbf{Level 3 (Hard Open-Ended Reasoning):}
Level 3 shows a stark contrast. GPT-4o achieves only 6.6\% accuracy with ICC=0.304, meaning 70\% of observed variance is trial-to-trial randomness rather than question difficulty. Single-run results are essentially unreliable. GPT-5 dramatically improves both accuracy (44.2\%, +37.6 pp) and consistency (ICC=0.629, +0.325). While still lower than Levels 1–2, GPT-5's higher ICC signals that this massive accuracy gain reflects genuine capability improvement, not lucky sampling. The improvement in ICC is particularly striking: it nearly doubles GPT-4o's value, suggesting GPT-5's superior reasoning makes hard questions succeed/fail more deterministically based on their actual difficulty. \\

\textbf{ICC Convergence Across GAIA Levels}
\\
\resizebox{0.45\textwidth}{!}{
\begin{tabular}{c|cc|cc|cc}
\hline
Trials & \multicolumn{2}{c|}{Level 1} & \multicolumn{2}{c|}{Level 2} & \multicolumn{2}{c}{Level 3} \\
(n) & GPT-4o & GPT-5 & GPT-4o & GPT-5 & GPT-4o & GPT-5 \\
\hline
$n = 2$ & 0.648 & 0.797 & 0.706 & 0.774 & 0.428 & 0.686 \\
$n = 4$ & 0.614 & 0.781 & 0.689 & 0.759 & 0.382 & 0.658 \\
$n = 8$ & 0.587 & 0.779 & 0.675 & 0.754 & 0.341 & 0.636 \\
$n = 16$ & 0.574 & 0.775 & 0.668 & 0.749 & 0.322 & 0.631 \\
$n = 32$ & 0.567 & 0.776 & 0.665 & 0.745 & 0.313 & 0.631 \\
$n = 64$ & 0.561 & 0.774 & 0.662 & 0.745 & 0.304 & 0.629 \\
\hline
\end{tabular}
}
\begin{center}
{\tiny $^*$Both GPT-4o and GPT-5 were evaluated with web search enabled.}
\end{center}

ICC stabilizes as more trials are collected. The decrease from $n=2$ to $n=64$ reflects regression to the mean: initial estimates with few trials overestimate between-question variance. As trials accumulate, more accurate estimates of true question difficulty emerge. \textbf{Convergence occurs by $n\approx 8-16$ for Levels 1--2 and by $n\approx 32$ for Level 3}, providing practitioners with a data-driven signal for setting resampling budgets. Level 3's slower convergence underscores its higher sensitivity to sample size. For a detailed model performance on GAIA refer to Appendix D.

\subsection{FRAMES: ICC in Retrieval and Factuality Tasks}
To validate that ICC reflects task structure beyond GAIA's open-ended reasoning, we evaluated on the FRAMES benchmark (tool-based web search and information retrieval. Due to computational cost concerns, we only took 50 questions, sampled randomly with a seed of 42, but still performed 64 trials per question. Table \ref{tab:frames_table} shows results.

FRAMES exhibits substantially higher ICC (ICC=0.7118) than GAIA Level 3 (ICC=0.304) and comparable ICC to GAIA Level 2 (ICC=0.662) (for GPT-4o). This emphasizes how the nature of the task can affect the reliability of results. However, GPT-5 with web search exhibits lower ICC (0.4955) than GPT-4o with web search (0.7355) on FRAMES despite having higher accuracy, suggesting agent capability can also introduce variability independent of task design.  This narrative emphasizes the importance of including ICC as an additional dimension of reporting in benchmark results. Though GPT-5 search has the highest accuracy, it also has the lowest stability and determinism in response when compared to GPT-4o and GPT-4o search.

\subsection{Deep Research Agents}
We evaluated o4-mini deep research on both benchmarks using only 8 trials per question due to computational cost. Interestingly, ICC values with only 8 trials (0.62-0.66) are comparable to or exceed many values in the main analysis despite the reduced sample size. 

\begin{enumerate}
    \item Frames (n=8, 100 questions): 74.6\% accuracy, ICC=0.664
    \item GAIA (n=8, full validation set, 165 questions): 40.0\% accuracy, ICC=0.621
\end{enumerate}

This could reflect several factors: (1) deep research's more deliberate reasoning may reduce trial-to-trial variability, (2) the larger question sets naturally stabilize ICC estimates, or (3) small-sample ICC can be unstable. However, with only a single deep research agent evaluated, we cannot draw firm conclusions. Larger-scale evaluation across multiple deep research agents would be needed to determine whether ICC patterns hold across different agent architectures and whether deep reasoning fundamentally reduces stochasticity.

\section{Implications for Agentic System Design}

\textbf{Sub-agent replacement decisions require both capability and consistency metrics.} When LLM-based agents become sub-components of larger systems, system-level reliability depends on both capability and consistency. A more capable sub-agent with low ICC introduces brittleness: downstream components cannot reliably predict behavior. Our framework shows that capability improvements without ICC improvements may not be robust under deployment. GAIA Level 3 demonstrates this: GPT-5's 7× accuracy improvement paired with doubled ICC signals a genuine, deployable improvement. Without ICC visibility, practitioners risk discovering inconsistency problems in production.

\textbf{Consistency is a tunable property independent of accuracy.} Beyond replacement decisions, ICC enables a new tuning target: consistency. Agent optimization typically focuses on accuracy. Yet it's possible to reduce within-question variance (improving ICC) through better prompting or tool design without changing accuracy. Such improvements are invisible to standard metrics but critical for system reliability. By monitoring ICC, practitioners directly optimize for the emergent properties—predictability and stability—that enable robust agentic systems.

To operationalize these insights, we propose the following reporting requirements and updated Evaluation Cards.
\\ \\
\textbf{Reporting Requirements}
Rather than report accuracy alone, practitioners should report:
\begin{equation}
 \text{Accuracy} \pm \text{95\% CI} \quad | \quad \text{ICC} \quad | \quad \text{Between-query SE}   
\end{equation}

This enables readers to understand (1) point estimate and uncertainty, (2) evaluation stability, and (3) agent consistency.
\\ \\
\textbf{Evaluation Cards}
We propose updated Evaluation Cards that capture run-level metadata: \\

\begin{table}[h]
\centering
\begin{tabular}{l|p{4cm}}
    \textbf{Field} & \textbf{Description} \\
    \hline
    Benchmark & Name and version (e.g., GAIA v2024 Level 2) \\
    Agent & Model name, version, decoding parameters (temperature, tools, etc.) \\
    Trials \& seeds & Number of trials, seed generation method \\
    Metrics & Mean accuracy with 95\% CI, ICC (variant reported), between-query SE \\
    Task complexity level & Level/difficulty designation (affects variance structure) \\
    Scoring details & Exact string match, fuzzy matching, normalization rules \\
    Limitations & Known caveats (e.g., highly stochastic level, small sample size) \\
\end{tabular}
\caption{Evaluation Cards: run-level metadata for agentic evaluation.}
\label{tab:evaluation_cards}
\end{table}
Together, these practices: reporting consistency metrics, documenting variance structure, and standardizing metadata, enable practitioners to build agentic systems with predictable, reliable sub-components. By making evaluation stability visible, we move toward principled composition of agentic systems where reliability is not hidden in leaderboard competition, but is explicit and measurable.

\section{Limitations and Future Work}

Scoring assumption: ICC analysis assumes binary (correct/incorrect) 
scoring; extension to partial-credit metrics requires reformulation. We use 
o4-mini as LLM judge without formal validation of inter-trial consistency; 
scorer variance may contribute to measured ICC but is expected to be minimal.
ICC variant choice:  High variance may reflect beneficial exploration 
in some domains; ICC measures consistency but does not prescribe optimal 
stochasticity levels. Our ICC(1,1) variant treats questions as random effects; 
alternative formulations may suit other contexts.
Generalization: Analysis focuses on reasoning and search tasks with 
OpenAI models (GPT-4o, GPT-5, o4-mini). ICC patterns may differ across other 
model families, embodied agents, or adversarial environments. FRAMES uses only 
50/824 questions due to computational constraints; GAIA uses the full validation 
set (165 questions). Deep research evaluation used one agent with 8 trials. Further research is needed for generalizable conclusions.
\section{Ethical Statement}
This work promotes evaluation transparency in agentic systems, enabling more informed deployment decisions about sub-agent reliability. Transparency in evaluation practices benefits the broader community by reducing opaque leaderboard competition and enabling practitioners to make principled choices about system composition.
However, we caution against misuse. High accuracy with low ICC indicates unreliable evaluation, not a suitable system for deployment—it signals the need for further investigation before deployment in safety-critical contexts. Practitioners should not use ICC metrics as justification to deploy agents with low consistency guarantees in high-stakes applications (healthcare, finance, autonomous systems).
Additionally, our analysis is limited to English-language benchmarks evaluated on specific model families. ICC patterns may differ across languages, domains, and agent architectures. We encourage evaluation rigor in under-studied settings, particularly for non-English and lower-resource applications where evaluation practices are less mature.

\section{Conclusion}
We introduced ICC as a metric for characterizing and reporting agentic evaluation stability. By analyzing GAIA across all three difficulty levels and FRAMES with 64 trials per question, we demonstrated that ICC varies dramatically with task structure: retrieval and reasoning tasks (Frames, $ICC\approx0.70$) exhibit clean evaluation, while reasoning questions testing multiple agentic capabilities (GAIA Level 3, ICC=0.30 for GPT-4o) are heavily influenced by agent stochasticity. This highlights a critical insight: task complexity affects not just absolute performance but evaluation reliability itself.
For practitioners building agentic systems, this distinction is essential. Accuracy improvements without ICC given as context may not comprehensively reflect genuine capability gains. We recommend reporting accuracy alongside ICC and within-query variance as standard practice, and propose updated Evaluation Cards to standardize this reporting. These simple practices: making variance visible and quantifying evaluation stability, enable practitioners to build systems with reliable, predictable sub-agents.
By moving agentic evaluation from opaque leaderboard competition toward principled, reproducible experimental science, we aim to establish evaluation rigor as a foundation for trustworthy agentic systems development.

\bibliography{aaai2026}

\clearpage
\onecolumn  
\appendix

\begin{center}
{\Large\bfseries Appendix A}
\end{center}
\label{app:appendixB}
\vspace{1em}

\begin{figure}[H]
\centering
\begin{minipage}{0.48\textwidth}
    \centering
    \includegraphics[width=\textwidth]{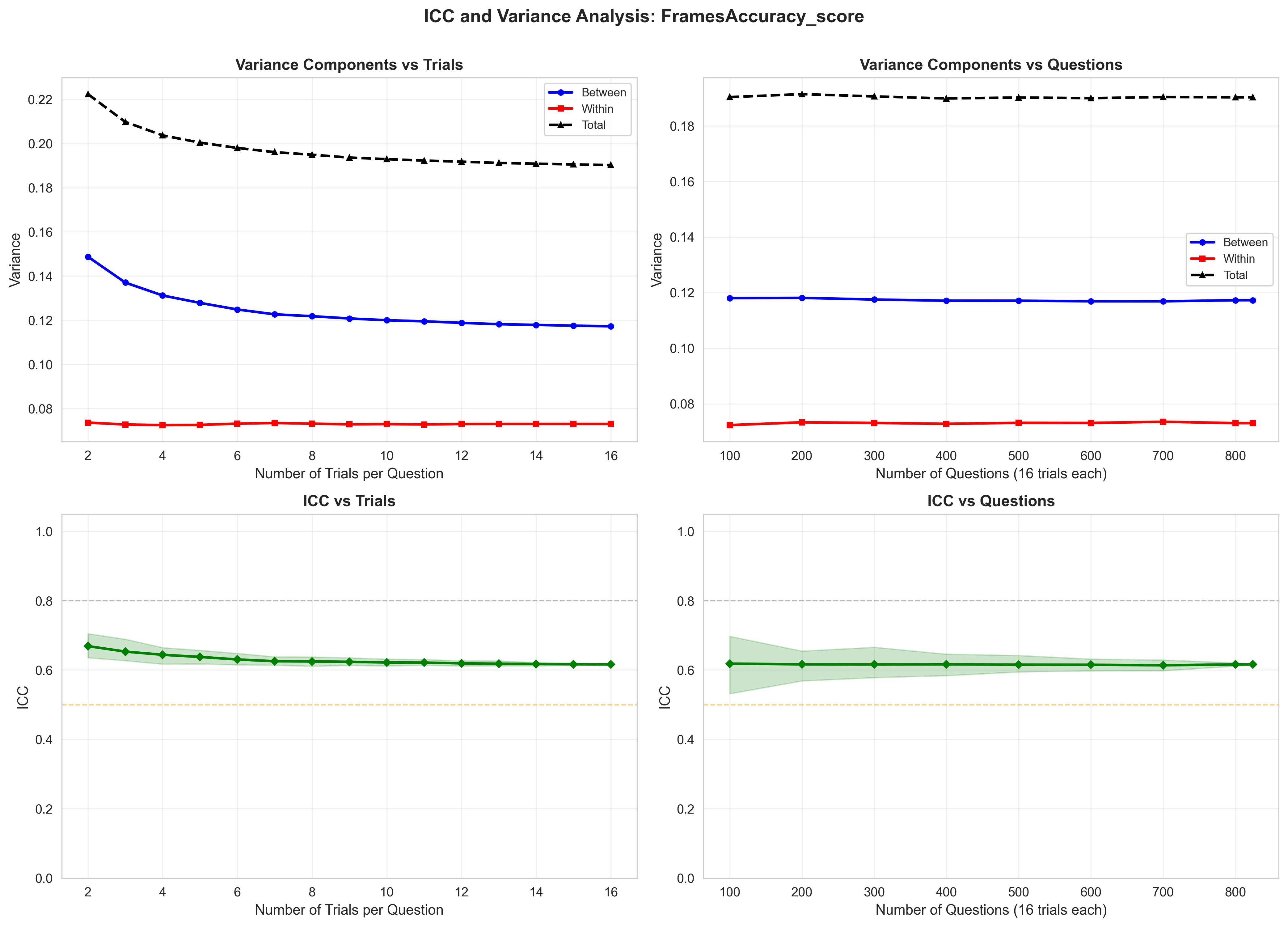}
    \caption{ICC Convergence Plot for FRAMES, GPT-5 Search.}
    \label{fig:icc_convergence_frames}
\end{minipage}
\hfill
\begin{minipage}{0.48\textwidth}
    \centering
    \includegraphics[width=\textwidth]{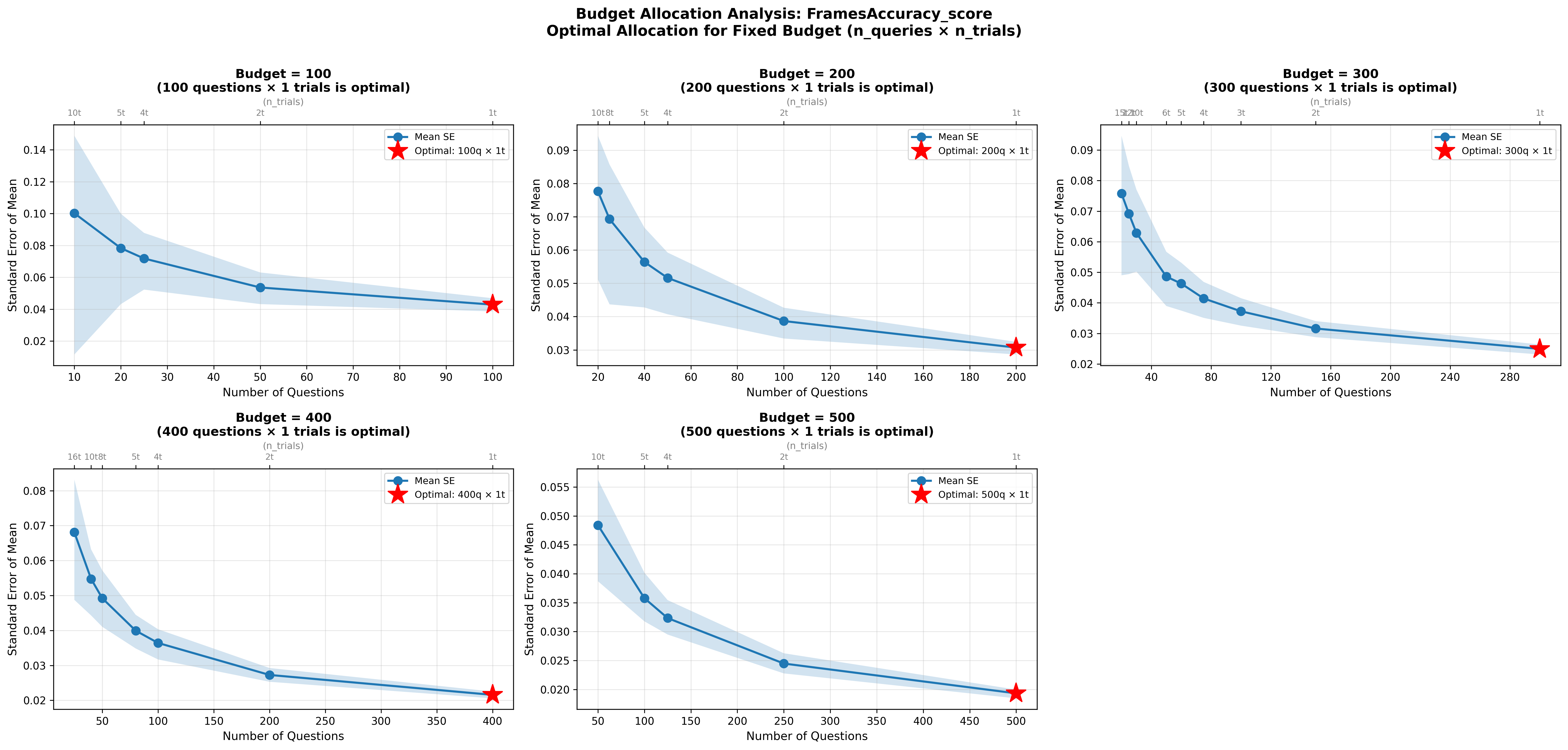}
    \caption{Budget Allocation Analysis.}
    \label{fig:variance_comparison}
\end{minipage}
\end{figure}
\begin{figure}[H]
\centering
\begin{minipage}{0.48\textwidth}
    \centering
    \includegraphics[width=\textwidth]{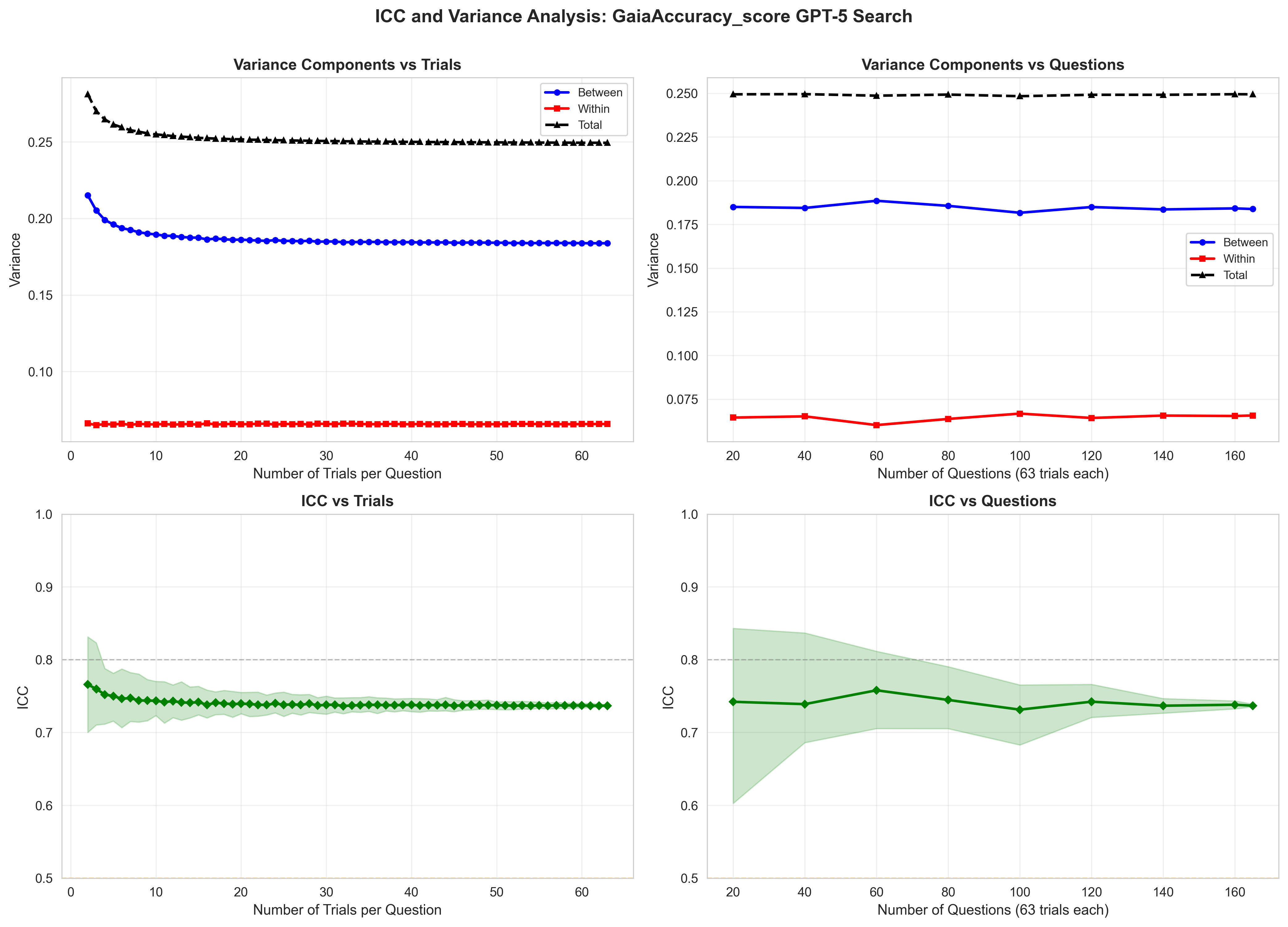}
    \caption{ICC Convergence Plot for GAIA, GPT-5 Search.}
    \label{fig:icc_gaia_gpt5}
\end{minipage}
\hfill
\begin{minipage}{0.48\textwidth}
    \centering
    \includegraphics[width=\textwidth]{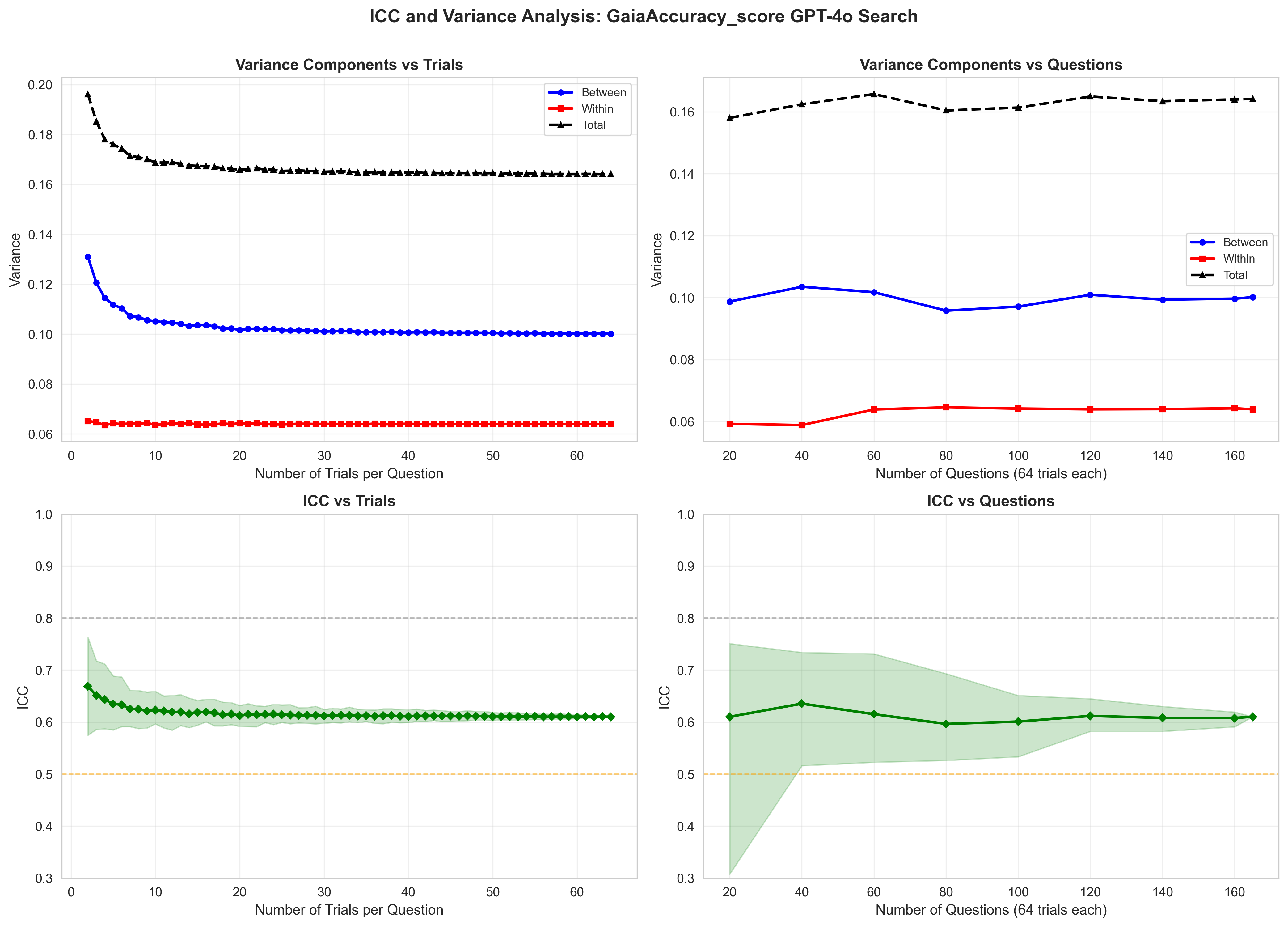}
    \caption{ICC Convergence Plot for GAIA, GPT-4o Search.}
    \label{fig:icc_gaia_gpt4o}
\end{minipage}
\end{figure}

\begin{center}
{\Large\bfseries Appendix B}
\end{center}
\vspace{1em}

\subsection*{Evaluation Prompt}
\label{prompt:evaluation}

\begin{quote}
\small\ttfamily
\noindent ===Task===\\
I need your help in evaluating an answer provided by an LLM against a ground truth
answer. Your task is to determine if the ground truth answer is present in the LLM's response.
Please analyze the provided data and make a decision.

\vspace{0.5em}
\noindent ===Instructions===\\
1. Carefully compare the "Predicted Answer" with the "Ground Truth Answer".\\
2. Consider the substance of the answers - look for equivalent information or correct answers. 
   Do not focus on exact wording unless the exact wording is crucial to the meaning.\\
3. Your final decision should be based on whether the meaning and the vital facts of the 
   "Ground Truth Answer" are present in the "Predicted Answer:"

\vspace{0.5em}
\noindent ===Input Data===\\
Question: \{question\}\\
Predicted Answer: \{predicted\_answer\}\\
Ground Truth Answer: \{target\}

\vspace{0.5em}
\noindent ===Output Format===\\
Provide your final evaluation in the following format:\\
"Explanation:" (How you made the decision?)\\
"Decision:" ("TRUE" or "FALSE")\\
Please proceed with the evaluation.
\end{quote}

\begin{center}
{\Large\bfseries Appendix C}
\end{center}
\vspace{1em}

\noindent All closed form models were evaluated via official APIs and open source models via fireworks.ai.
\begin{itemize}
    \item \textbf{GPT-4o search}: \texttt{gpt-4o-search-preview}
    \item \textbf{GPT-4o}: \texttt{gpt-4o}
    \item \textbf{GPT-5 search}: \texttt{gpt-5} with web search tool enabled
    \item \textbf{o4-mini deep research}: \texttt{o4-mini-deep-research}
    \item \textbf{o4-mini (judge)}: \texttt{o4-mini}
    \item \textbf{Claude 4.5 Sonnet}: \texttt{claude-sonnet-4-5}
    \item \textbf{Claude 4.5 Haiku}: \texttt{claude-haiku-4-5}
    \item \textbf{Gemini 2.5 Pro}: \texttt{gemini-2.5-pro} 
     \item \textbf{Qwen3-235b-a22b}: \texttt{qwen3-235b-a22b} 
      \item \textbf{Deepseek-v3p1}: \texttt{deepseek-v3p1} 
    
\end{itemize}

\vspace{0.5em}
\noindent Evaluation conducted: October 2025 \\ \\
API Documentation: 
\begin{itemize}
    \item \url{https://platform.openai.com/docs/models}
    \item \url{https://docs.claude.com/en/docs/about-claude/models/overview}
     \item \url{https://ai.google.dev/gemini-api/docs/models}
     \item \url{https://docs.fireworks.ai/api-reference}

\end{itemize}

\begin{center} {\Large\bfseries Appendix D} \end{center} \vspace{1em} 
\small
\begin{table*}[h]
    \centering
\begin{tabular}{c|c|c|c|c}
\textbf{Model} & \textbf{Accuracy} & \textbf{95\% CI} & \textbf{Between Var} & \textbf{ICC} \\
\hline

    GPT-5 search & 59.44\% & [47.28\%, 71.60\%] & 0.183 & \textbf{0.745} \\
    GPT-4o search & 21.59\% & [11.92\%, 31.27\%] & 0.115 & \textbf{0.671} \\
    
Claude 4.5 Sonnet& {39.71\%}& {[27.50\%, 51.91\%]}& {0.184}& \textbf{0.756}\\
Claude 4.5 Haiku& {32.26\%}& {[20.76\%, 43.76\%]}& {0.164}& \textbf{0.738}\\

Gemini 2.5 Pro & 31.28\% & [19.65\%, 42.92\%] & 0.180 & \textbf{0.772}\\
  Qwen3-235b-a22b & 12.97\% & [4.70\%, 21.24\%] & 0.085 & \textbf{0.736}\\
 Deepseek-v3p1 & 22.28\% & [12.30\%, 32.26\%] & 0.123 & \textbf{0.699}\\
\end{tabular}
\begin{center}
{\tiny $^*$OpenAI \& Claude family were evaluated with web search and others without web search.}
\end{center}
\begin{center}
\caption{ICC and variance decomposition on GAIA (n=50 with random seed of 42, 64 trials per question).}
\end{center}
\label{tab:gaia_all_model_families}
\end{table*}
Across seven models, we show a critical tension in agentic AI trustworthiness: capability vs consistency. GPT-5 search achieves the highest accuracy (59.44\%) but exhibits lower ICC (0.745) than Claude 4.5 Sonnet (ICC 0.756 at 39.71\% accuracy), indicating that performance gains introduce unpredictability. Other (Deepseek, Qwen, Gemini) degrade gracefully on harder tasks, maintaining modest ICC (0.70–0.77), while Claude models cluster consistently around ICC 0.73–0.77 regardless of accuracy variation. These results indicate that accuracy and ICC should be jointly reported when evaluating agentic systems. Reporting accuracy alone provides incomplete information for system reliability assessment. We recommend ICC and query variance as standard evaluation metrics alongside accuracy to enable practitioners to make trustworthy decisions about agent selection and deployment.
\clearpage
\begin{center}
{\Large\bfseries Appendix E}
\end{center}
\label{app:appendixA}
\vspace{1em}
\subsubsection{Intraclass Correlation Coefficient (ICC) Mathematical Framework}

Let $Y_{ij}$ denote the performance score of the $j$-th trial on the $i$-th task, where $i = 1, 2, \ldots, n$ tasks and $j = 1, 2, \ldots, T_i$ trials per task. Under a one-way random effects model:

\begin{equation}
Y_{ij} = \mu + \alpha_i + \varepsilon_{ij}
\label{eq:random_effects}
\end{equation}

where $\alpha_i \sim N(0, \sigma_\alpha^2)$ represents the random task effect and $\varepsilon_{ij} \sim N(0, \sigma_\varepsilon^2)$ represents the within-task error term.

The within-task variance, representing trial-to-trial inconsistency, is computed as:

\begin{equation}
\sigma_w^2 = \frac{\sum_{i=1}^{n} \sum_{j=1}^{T_i} (Y_{ij} - \bar{Y}_{i\cdot})^2}{\sum_{i=1}^{n} (T_i - 1)}
\label{eq:within_variance}
\end{equation}

where $\bar{Y}_{i\cdot} = \frac{1}{T_i}\sum_{j=1}^{T_i} Y_{ij}$ is the mean performance on task $i$. This can equivalently be expressed as a weighted average of individual task variances:

\begin{equation}
\sigma_w^2 = \frac{\sum_{i=1}^{n} (T_i - 1) s_i^2}{\sum_{i=1}^{n} (T_i - 1)}
\label{eq:pooled_within}
\end{equation}

where $s_i^2 = \frac{1}{T_i-1}\sum_{j=1}^{T_i} (Y_{ij} - \bar{Y}_{i\cdot})^2$ is the sample variance of task $i$. Weighting by degrees of freedom $(T_i - 1)$ ensures that tasks with more trials contribute proportionally more to the pooled variance estimate, providing an unbiased estimator when trial counts vary between tasks.

When the number of trials is equal across all tasks ($T_i = T$ for all $i$), this weighted average reduces to the arithmetic mean:

\begin{equation}
\sigma_w^2 = \frac{1}{n} \sum_{i=1}^{n} s_i^2
\label{eq:simple_mean_within}
\end{equation}
\clearpage

\end{document}